\documentclass[letterpaper, 10 pt, conference]{ieeeconf}  
\usepackage[utf8]{inputenc}
\usepackage{cite}
\usepackage[hidelinks]{hyperref}
\usepackage{todonotes}
\usepackage{graphicx}
\usepackage{xcolor}
\usepackage{booktabs}
\usepackage{multirow}
\usepackage{siunitx}

\renewcommand{\paragraph}[1]{%
    \vspace{0.5em}\noindent\textbf{#1}\quad
}
\IEEEoverridecommandlockouts                              

\title{\LARGE \bf
GlowTact: Simple and Compact Vision-Based Tactile Sensing with High Sensitivity and Spatial Resolution
}
\author{
Yuxiang Ma$^{1,*}$,
Megha Tippur$^{1,*}$,
Pengfei Ye$^{1}$,
Sandra Q. Liu$^{1}$,
Haonan Chen$^{1,2}$,\\
Francis Richard Cottrell$^{1}$,
Edward Adelson$^{1}$ \\
$^{1}$Massachusetts Institute of Technology, Cambridge, MA, USA \\
$^{2}$Harvard University, Cambridge, MA, USA \\
$^{*}$Equal contribution 
}

\begin{document}

\maketitle
\thispagestyle{empty}
\pagestyle{empty}

\begin{abstract}

Vision-based tactile sensors (VBTS) provide rich contact information for robotic manipulation, but existing designs can be hard to simplify and adapt to the size and constraints of humanoid fingertips. We introduce \textbf{GlowTact}, a pressure-responsive vision-based tactile sensing mechanism that directly visualizes contact pressure. \textbf{GlowTact} requires only single-color, non-directional illumination, and the raw tactile image directly represents the pressure distribution without explicit geometry reconstruction. This simple sensing principle enables compact, customizable tactile sensors while preserving high sensitivity and rich spatial detail. We demonstrate gram-scale contact detection, accurate normal-force estimation, and reconstruction of fine contact geometry, including M1 screw threads. These results establish \textbf{GlowTact}  as a practical new sensing technology for compact humanoid fingertips, combining a durable nitrile membrane and simple optical design with sensitive and information-rich tactile perception.

\end{abstract}

\section{INTRODUCTION}

Tactile sensing can reveal important information such as contact location, object
orientation, object identity, and object motion~\cite{yuan2017gelsight, dong2017improved,
taylor2022gelslim}. Vision-based tactile sensors (VBTS) are particularly
attractive because they provide dense, high-resolution contact
observations~\cite{johnson2009retrographic, yuan2017gelsight,
shimonomura2019tactile}.

\begin{figure}[!ht]
    \centering
    \includegraphics[width=\columnwidth]{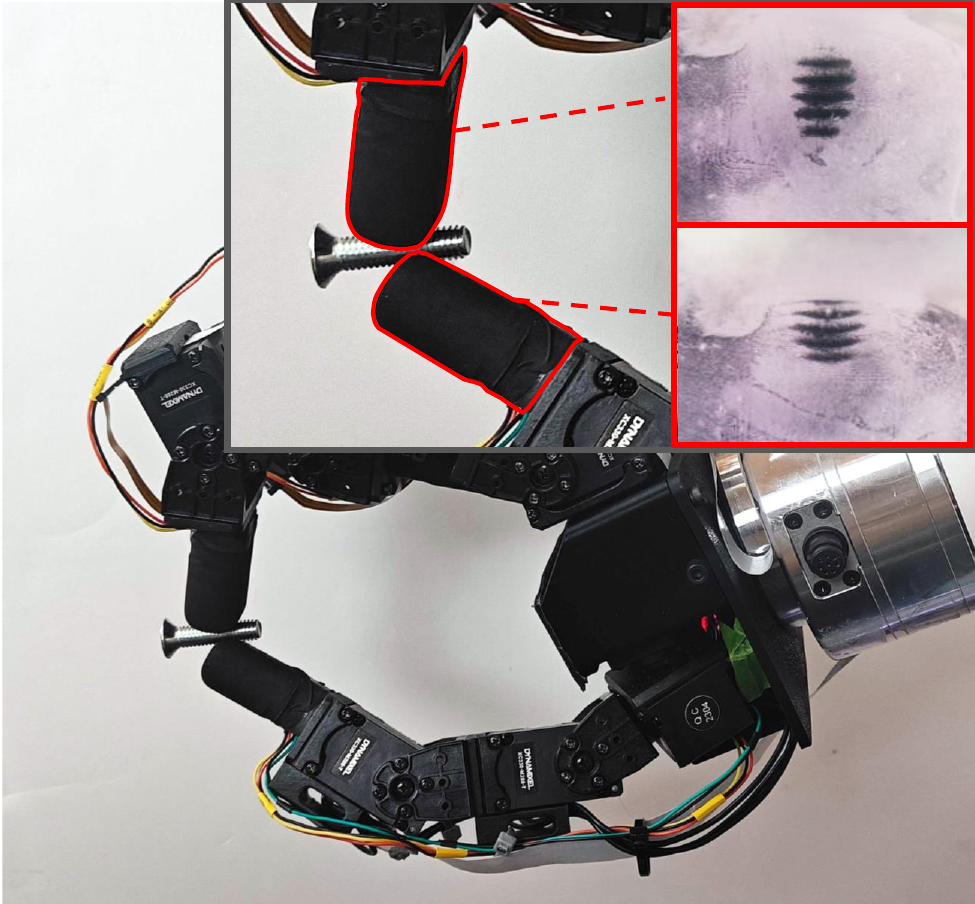}
    \caption{\textbf{GlowTact is a pressure-responsive vision-based tactile sensing mechanism}. A GlowTact-equipped LEAP hand grasps a screw (left), while the tactile images (right) directly visualize contact pressure and fine thread geometry. GlowTact combines sensitive low-force response with rich spatial detail using only single-color illumination. }
    \label{fig:teaser}
    \vspace{-10 pt}
\end{figure}
There are various methods to building VBTS, each with its pros and cons. These fabrication approaches include utilizing color versus monochrome cameras, having differing complexity of lighting systems, using various amounts of computational load, and they result in a wide range of sensor resolution and sensitivity. Our new sensor, GlowTact, scores well on all of these dimensions: it is monochrome, needs almost no computation, uses simple lighting, has high resolution and high sensitivity.

For humanoid fingers, minimal computation is desirable because each fingertip needs its own camera, along with the associated data stream and computation. Humanoid fingers can also raise challenges for multicolored lighting design due to the 3D shape of the finger. Moreover, most VBTS work by measuring the deformation of a gel pad, meaning that these sensors are unsuitable for delicate manipulation tasks because they are limited by gel sensitivity. GlowTact simplifies all of these challenges, making it especially suitable for use in humanoid fingers. 

The sensing principle is novel: Instead of coating the gel with an opaque
layer~\cite{yuan2017gelsight, donlon2018gelslim, lambeta2020digit,
lin20239dtact}, GlowTact uses a black elastomer membrane that sits unbonded on the microtextured surface of a clear
gel. A microscopic air gap keeps the membrane
optically uncoupled from the gel, so light is reflected at the gel-air
interface, and the camera images a uniform light gray surface in the default state. Contact presses the
membrane into the texture and causes a dark
region to appear at the contact point. The imaging principle is related to frustrated total internal reflection (FTIR), but most of the reflection is partial, not total, and it is diffuse. The air gap is tiny, so GlowTact can produce visible signals with small forces. Moreover, the membrane can be made of a robust material such as nitrile rubber, leading to a durable sensing surface. Additionally, the sensor requires no reflective coating, directional multicolor illumination, or photometric calibration, and can be adapted to form factors as small as a human fingertip.

We discuss GlowTact's  performance in a flat and rounded humanoid formats. We show that GlowTact directly signals pressure, mapping local pressure into local darkening and demonstrate its high resolution and high sensitivity to low forces.


%
%
%
%

\section{RELATED WORK}
Robotic tactile sensors employ a variety of transduction mechanisms, such as
resistive, capacitive, magnetic, acoustic, and barometric sensing, each
involving different tradeoffs among resolution, bandwidth, and form
factor~\cite{sundaram2019learning,fishel2012sensing,yong2022soft,tang2019design,zainuddin2015resistive,heyneman2012biologically,muhlbacher2015responsive,saloutos2023design,epstein2020bimodal,alspach2019soft,wardcherrier2018tactip,lambeta2020digit,shimonomura2019tactile}.
These options support applications at different scales, from tactile skins
that provide coarse contact detection across a humanoid's
body~\cite{mittendorfer2011humanoid,mittendorfer2015realizing,schmitz2011methods,cheng2019comprehensive,barreiros2025learning}
to individual fingertip taxels for fine
manipulation~\cite{fishel2012sensing,saloutos2023design}. VBTS have proven
especially effective for adding touch sensing to robotic end
effectors~\cite{johnson2009retrographic,yuan2017gelsight,taylor2022gelslim}.
Their cameras produce high-resolution images of the contact interface that can
encode surface geometry, object pose, pressure and force distributions, and
slip~\cite{yuan2017gelsight,dong2017improved,taylor2022gelslim,sun2022soft,lin20239dtact}.
Because their outputs are images, VBTS integrate directly with vision-based
deep-learning methods.

A number of VBTS designs have been introduced over the years. In general,
these sensors combine a soft elastomer, some form of internal illumination,
and a camera housed at the base of the sensor, so that when an object contacts the
elastomer, the resulting deformation is observed directly in the camera image
at very high spatial resolution \cite{johnson2009retrographic, yuan2017gelsight, zhang2018fingervision}. The design
space is largely defined by how that deformation is encoded optically and how
it is recovered computationally.

\subsection{Vision-Based Tactile Sensors}
The dominant approach in robotic VBTS, Ge;Sight, recovers surface geometry from shading.
\cite{johnson2009retrographic} placed an opaque, diffusely reflecting membrane on a clear elastomer and illuminated it from several
directions so that photometric stereo could recover surface gradients that are
then integrated into a height map. Yuan et al. adapted this technology to a
robotic form factor with the sensor introduced in \cite{yuan2017gelsight}. A
large family of sensors has since followed, targeting higher reconstruction fidelity
and smaller packaging \cite{wang2021gelsight, dong2017improved, taylor2022gelslim}.  As with other VBTS, GelSight sensors can measure shear displacement if an array of markers is printed on the inner membrane and tracked over time. A second VBTS family encodes
deformation through the motion of physical features instead of shading. Markers
printed on or embedded beneath the elastomer membrane are tracked across frames,
producing a displacement field from which shear and slip can be inferred
\cite{wardcherrier2018tactip, zhang2018fingervision, taylor2022gelslim}. 

Other sensors recover surface geometry more directly. Soft-bubble sensors observe an inflated membrane with an
internal depth camera \cite{alspach2019soft}. The DTact \cite{lin2022dtact} and 9DTact \cite{lin20239dtact} sensors exploit light attenuation, where a
translucent gel beneath a black coating produces a haze whose local brightness
varies with gel thickness. These image encodings all
ultimately measure the deformed geometry of the
elastomer. Notably, contact pressure and normal force are never observed directly; instead, they
are inferred from that geometry, either through calibrated models of the
elastomer mechanics or through networks trained on probing data with force
labels. 

\section{GlowTact Sensing Principle and Design}

\begin{figure*}[!ht]
    \centering
    \includegraphics[width=\textwidth]{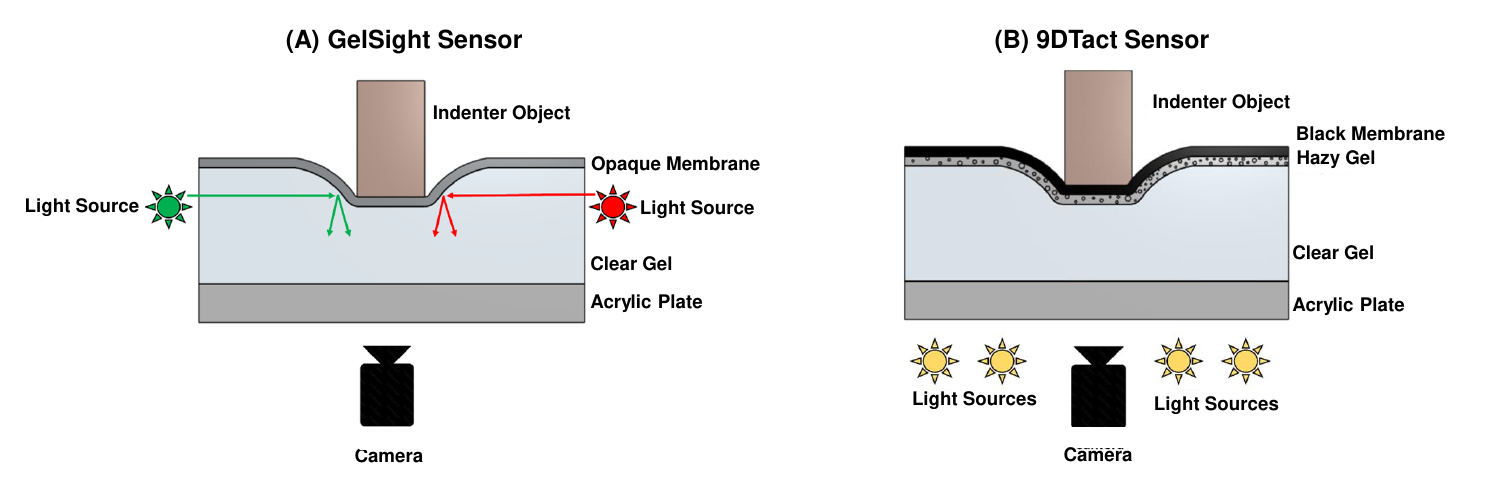}
    \vspace{-30 pt}
    \caption{\textbf{Working principles of representative vision-based tactile
    sensors.} \textbf{(A)} GelSight uses directional illumination and an opaque
    reflective membrane to encode surface geometry through shading.
    \textbf{(B)} 9DTact uses light attenuation in a translucent gel beneath a
    black membrane to encode local gel thickness in image intensity.}
    \label{comparison_other_sensors}
    \vspace{-10 pt}
\end{figure*}

\begin{figure*}[!ht]
    \centering
    \includegraphics[width=0.77\textwidth]{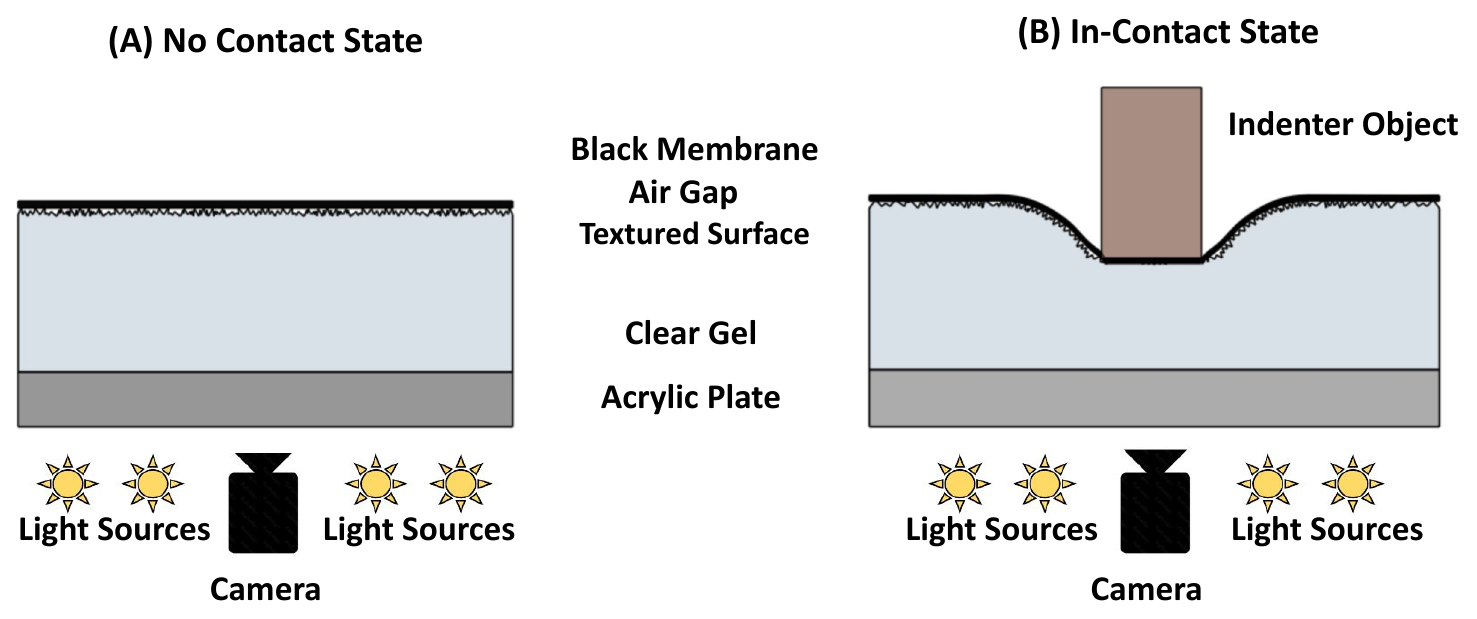}
    \vspace{-10 pt}
    \caption{\textbf{GlowTact sensing principle.} In the unloaded state \textbf{(A)}, a thin air gap separates the black membrane from the gel's micro-textured surface, so light within the gel is diffusely reflected at the gel–air interface, and the camera sees a uniform light gray. When an indenter presses the membrane into optical contact with the gel \textbf{(B)}, the reflective interface is eliminated and light is instead absorbed by the black membrane, darkening the contact region as pressure increases.}
    \label{glowtact_technology}
\end{figure*}

\subsection{GlowTact Design Principles}
Many of the prior vision-based tactile sensor designs~\cite{yuan2017gelsight, donlon2018gelslim, wang2021gelsight, lambeta2020digit, lin20239dtact} consist of a clear elastomeric gel pad mounted on a rigid, transparent acrylic backing. When an object comes in contact with the sensor's surface, the compliant gel pad changes shape as a result of the applied forces. A camera housed at the base of the sensor observes the gel surface through the clear substrates, while LEDs provide the internal illumination needed to capture deformations of the elastomer~\cite{tippur2024rainbowsight, tippur2023gelsight360}.

In the case of GelSight-based sensors, red, green, and blue LEDs illuminate the gel from three distinct directions, and an opaque layer of gray or metallic elastomeric paint is applied to the top surface of the gel. By analyzing the independent shading produced by each illumination direction on its corresponding color channel, photometric stereo can be used to estimate the surface gradients of the deformed gel. These gradients can then be integrated to produce a depth or height map of the contact surface. \autoref{comparison_other_sensors} illustrates the GelSight technology. Alternatively, the DTact and 9DTact sensor designs introduced in~\cite{lin2022dtact, lin20239dtact} offer another VBTS design approach. Here, the gel pad is made of a translucent material covered with an opaque black elastomeric coating. When the pad is non-directionally illuminated from below, the camera image appears filled with a diffuse haze produced by light scattering within the translucent layer, and the camera measures the point-to-point brightness of this haze. As an object in contact with the sensor skin compresses the translucent layer, dimmer regions indicate locally thinner gel, allowing brightness to be converted directly into a height map from a single calibration image.

\begin{figure}[!ht]
    \centering
    \includegraphics[width=\columnwidth]{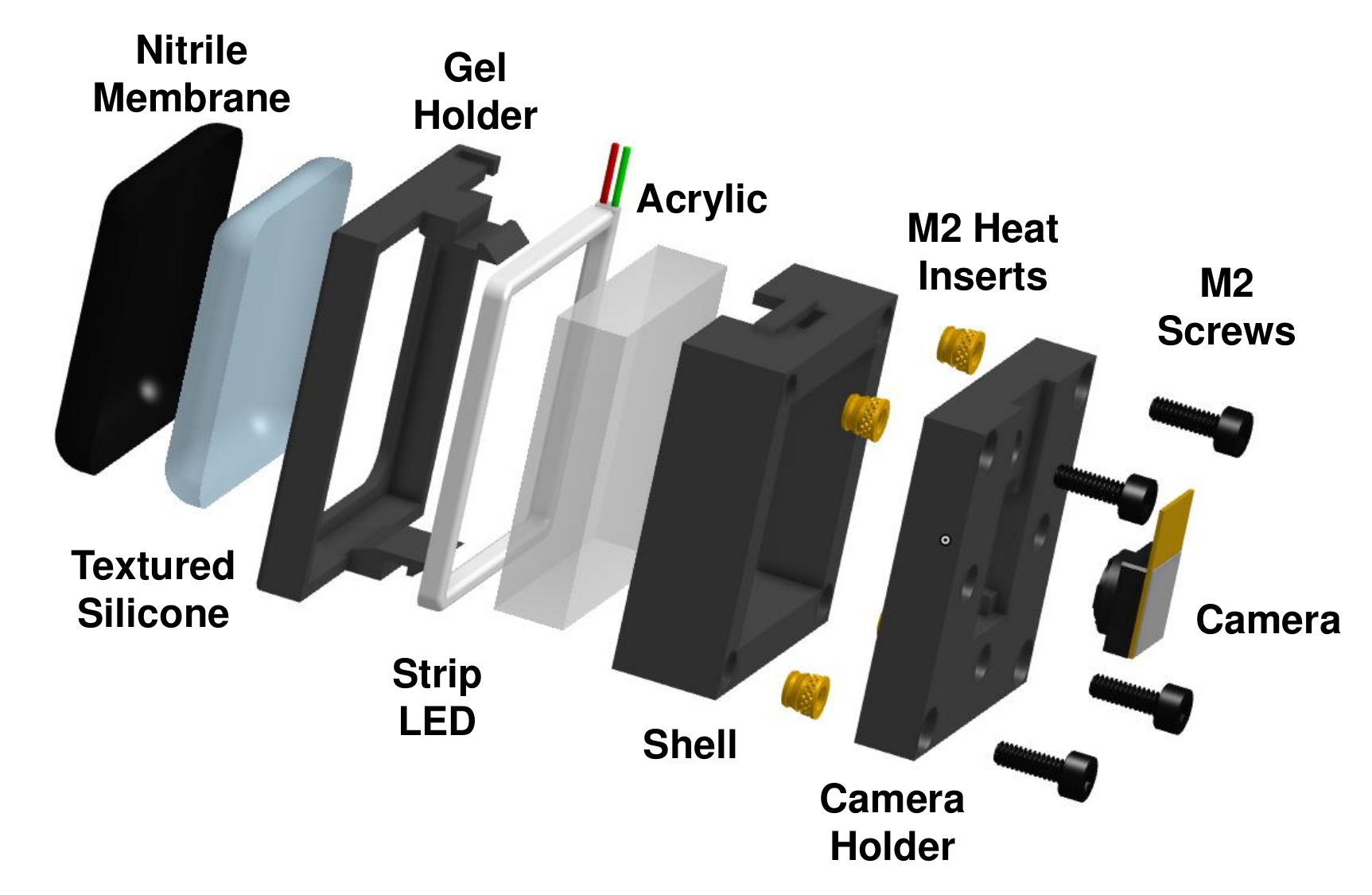}
    \caption{\textbf{Exploded view of the Flat GlowTact sensor.} A black membrane cut from a nitrile glove is stretched across the textured silicone gel pad, which is supported by the gel holder, forming the pressure-sensitive optical interface. A perimeter LED strip illuminates the interface through an acrylic plate, while a fisheye camera mounted in the camera holder observes it from within the sensor shell.}
    \label{flat_sensor_exploded}
\end{figure}

The GlowTact sensor introduced in this work differs from previous VBTS designs in that \textbf{it provides a direct estimate of pressure across the contact surface}. This is accomplished through an optical sensing principle that, to our knowledge, is novel among VBTS, and that additionally lends itself to simple design and fabrication methods. \autoref{flat_sensor_exploded} shows an exploded view of a flat GlowTact sensor. Similar to prior VBTS, a clear gel elastomer is mounted on a rigid acrylic plate. However, rather than coating the gel's outer surface with an opaque elastomeric paint as in~\cite{yuan2017gelsight, donlon2018gelslim}, the outer surface of the clear gel pad is fabricated with a \textbf{fine-grained, random texture}.  A thin black elastomeric membrane lies unbonded over the textured surface. Because the membrane contacts the gel only at the peaks of the microtexture, a thin air gap separates the two, leaving the membrane optically uncoupled from the gel. When light from within the gel pad strikes the rough gel-to-air interface, much of the light is diffusely reflected back toward the camera due to the refractive index mismatch between the gel and the air above. Therefore, when nothing is in contact with the sensor surface, the camera sees this diffuse reflection as a uniform light gray. However, when an indenter presses against the black membrane, it compresses the gel's texture and brings the black membrane into optical contact with the gel, eliminating the reflective interface; light instead passes into the black membrane and is absorbed, making the region appear dark. The greater the pressure, the more complete the optical coupling, and the darker the patch appears. Thus, in a GlowTact sensor, local darkness is a direct indication of local pressure.

\section{Sensor Fabrication and Implementations}
\label{sec:fabrication}
\subsection{Flat GlowTact Sensor}
\label{sec:flat_fab}
The flat GlowTact gel pad is fabricated using an open-mold casting process. A sheet of P2500 aluminum oxide sandpaper is adhered to the base of the mold with its abrasive surface facing upward.  Two thin coats of Inhibit X (Smooth-On Inc.) are applied onto the sandpaper to prevent cure inhibition. 

A 4 mm-thick layer of clear silicone gel (XP-565, Silicones Inc.), mixed at a 10:1 ratio by weight,  is degassed under vacuum and poured into the mold. The gel is cured at room temperature for 4 h, then post-cured at 50°C for 2 h in a dehydrator to complete crosslinking and eliminate residual surface tackiness. After curing, the gel is carefully de-molded. The resultant gel has a random microscale texture inherited from the sandpaper. The textured surface is cleaned with isopropyl alcohol (IPA) to remove residual mold contaminants. The gel pad is bonded to a 25.4 mm (1 in) square, 6.35 mm (0.25 in) thick clear PMMA acrylic plate using a silicone adhesive (Factor II A-564).

The sensor assembly is illustrated in \autoref{flat_sensor_exploded}. The contact membrane is cut from the wrist of a thin black nitrile glove. The square membrane is stretched taut across the front gel holder and secured around its inner lip using cyanoacrylate adhesive. The gel assembly is press-fit into the gel holder such that in the unloaded state, the membrane lightly contacts only the textured silicone surface asperity peaks.

The acrylic plate serves as the gel mechanical backing and optical light guide for illumination. A single-color LED strip mounted around the acrylic perimeter provides approximately uniform internal illumination, while a fisheye camera positioned beneath the acrylic captures the tactile response.

\begin{figure}[!ht]
    \centering
    \includegraphics[width=\columnwidth]{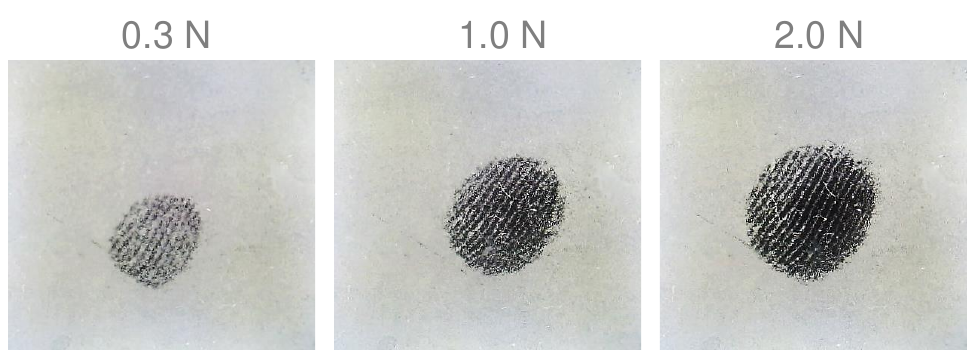}
    \caption{\textbf{Response of flat GlowTact sensor.} Fingerprint images demonstrate high sensitivity and high resolution, shown for normal loads of 0.3 N, 1.0 N, and 2.0 N. Increasing force results in progressively stronger optical coupling and a larger contact area while preserving fine ridge patterns.}
    \label{fingerprints}
\end{figure}

\autoref{fingerprints} illustrates the raw image output of a flat GlowTact sensor as a human finger is pressed into it. The images show that as the finger presses with greater force, the contact becomes darker. This direct signal means that GlowTact does not need as much image processing as other VBTS do to interpret contact information. Furthermore, all images preserve the fingerprint ridges. Thus, the GlowTact image achieves high sensitivity and resolution, while it also directly signals contact pressure. 

\begin{figure}[!ht]
    \centering
    \includegraphics[width=\columnwidth]{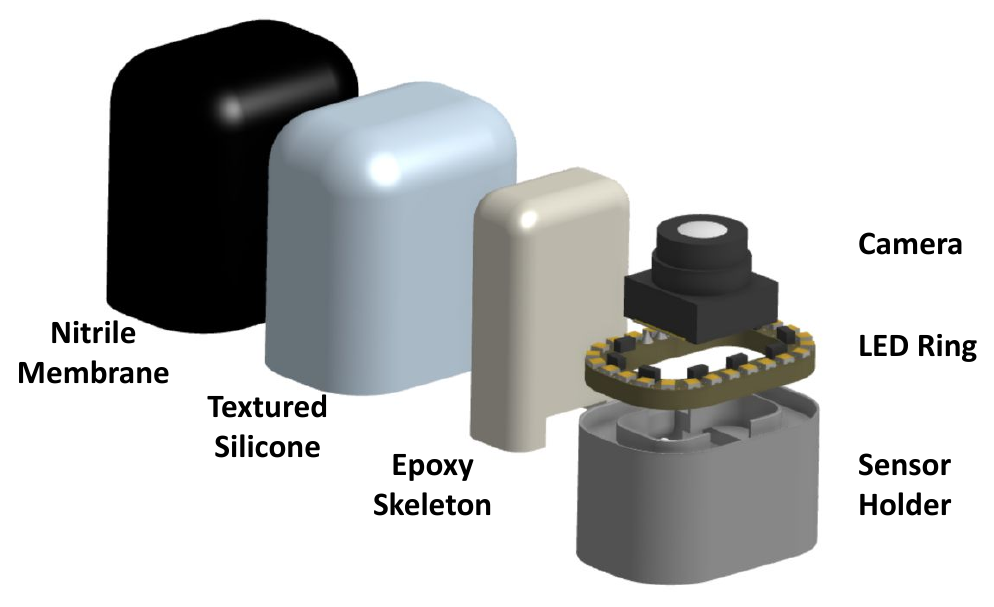}
    \caption{\textbf{Exploded view of the Omnidirectional GlowTact fingertip sensor.} A black membrane cut from a nitrile glove is stretched across the surface of the textured silicone, which coats a clear epoxy skeleton. A single-color LED ring and fisheye camera sit at the base of the sensor.}
    \label{fingertip_sensor_exploded}
\end{figure}

\subsection{Fingertip GlowTact Sensors}

To demonstrate compatibility with compact curved geometries, we developed two fingertip implementations: an omnidirectional sensor with tactile coverage around the fingertip and a humanoid-scale sensor with front-facing coverage. The two designs use the same pressure-induced optical-coupling principle but employ different methods for forming the textured curved surface.

\paragraph{Omnidirectional Fingertip Sensor}
The components of a fingertip design are shown in \autoref{fingertip_sensor_exploded}, adapted from the design and manufacturing process introduced in~\cite{tippur2024rainbowsight}, specifically of the internal skeleton. However, because the outer surface of a GlowTact elastomer must have the microtexture, a new mold fabrication process is required.

One challenge of mold fabrication for non-flat fingertips is that the surface microgeometry textures cannot be directly 3D printed on common SLA resin printers. To resolve this issue, we first print a convex mold negative of the sensor shape (Formlabs Form 3, Tough 2000 resin). The surface of the negative is rough-sanded to remove 3D printing layer lines and to promote adhesion in the subsequent step. After the negative is wiped down with IPA, a liquid silicone medical adhesive (Skinster Medical Adhesive) is evenly stippled across its surface. This application is repeated two times, with a 5 minute wait time to prevent dripping. 

To add the texture to the surface of the negative, the piece is fully submerged in a container of \SI{9}{\micro\meter} aluminum oxide powder. Next, compressed air is used to blow off excess powder. This powder-coating step is repeated two more times to ensure the entire sensing surface is uniformly coated. Finally, a light layer of mold release is sprayed onto the mold negative, and a concave silicone mold is cast against it using Smooth-On Mold Star 20T.

The inner surface of the silicone mold is sprayed with a layer of mold release (Mann Ease Release 200), and clear silicone (XP-565, mixed at a 10:1 base-to-activator ratio by weight) is poured into the mold. The rigid internal skeleton of the sensor is submerged into the filled cavity, and the assembly is cured following the same procedure described in Section~\ref{sec:flat_fab}. We note that the intermediate silicone mold is necessary, as applying the medical adhesive directly to the inside of a 3D-printed concave mold causes cure inhibition of the XP-565 elastomer, leaving an unwanted tacky finish on the final sensors.

Similar to the flat sensor, the thin, black elastomeric skin for the fingertip GlowTact sensor is created by cutting the fingertip from a black nitrile glove. The glove fingertip is stretched over the sensor surface and secured to the sensor holder with glue or double-sided tape. The glove size is selected according to the sensor dimensions, ensuring the skin stretches tautly over the surface without excessive tension. 

\begin{figure}
    \centering
    \includegraphics[width=\columnwidth]{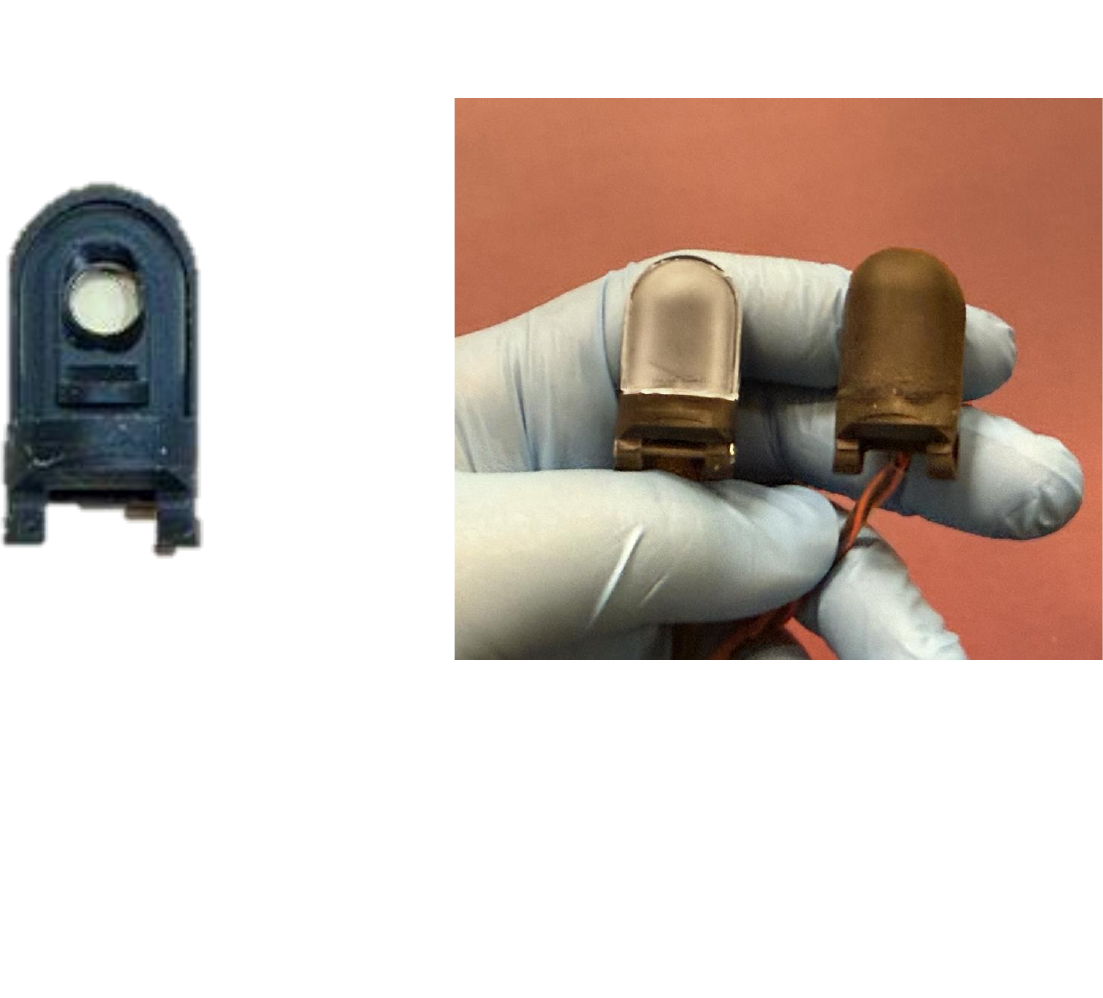}
    \caption{
    \textbf{Humanoid GlowTact fingertip sensor.}
    (a) Main components of the sensor: wide-angle camera, U-shaped LED strip,
    finger housing, and curved gel sensing body.
    (b) Assembled fingertip without and with the black nitrile contact membrane.
    (c) Raw tactile image produced by pressing a U.S. quarter against the sensor;
    dark regions indicate increased local pressure, while the coin edge and raised
    lettering remain visible.
    }
    \label{fig:humanoid_finger}
\end{figure}

\paragraph{Humanoid Fingertip Sensor}
We also fabricated a rounded GlowTact sensor, shown in \autoref{fig:humanoid_finger}. This sensor is about the shape and size of an adult human fingertip, at 17 mm wide. Rather than the full circumference of the finger, the sensing region covers only the front portion. The components are shown in \autoref{fig:humanoid_finger} a. A small wide angle camera is aimed perpendicular to the sensor. A U-shaped strip of LEDs illuminates the sensor along its rim. The sensor itself is a flattened hemicylinder capped with a quadrant of a sphere. A 3 mm layer of clear gel (XP-565, 15:1, Silicones Inc.), mounted on a rigid support, forms the body of the sensor. The gel is covered with a layer of microtextured Smooth-on MoldStar 20T, which is created from casting a 100 microns thick film on top of P2500 aluminum oxide sandpaper. This film is stretched over the rounded sensor, and glued in place with A-564 silicone sealant (Factor II).

The black skin membrane is made from cutting the pinky finger of a nitrile glove, although using standard black latex finger cots yields similar results. As with the flat GlowTact sensor, the humanoid-sclae GlowTact fingertip sensor has high sensitivity and resolution, and can see the raised lettering on a U.S. quarter coin (\autoref{fig:humanoid_finger}).

\section{EXPERIMENTAL EVALUATION}

Using a basic flat sensor, we evaluate GlowTact from three different perspectives:

\noindent\textbf{Contact geometry.} Can GlowTact recover spatial contact information despite relying on pressure-induced optical response? We assess this qualitatively through three-dimensional contact-shape reconstruction.

\noindent\textbf{Contact Sensitivity.} How effectively does GlowTact detect contact and weak normal forces? We compare its signal-to-noise ratio and low-force contact-detection performance with the GelSight Mini~\cite{gelsightmini}, a widely used commercial VBTS based on the photometric-stereo sensing principle of~\cite{yuan2017gelsight}.

\noindent\textbf{Force estimation.}  Does the GlowTact signal support
accurate normal-force estimation across probe geometries and everyday
objects? We evaluate learned models on spatially held-out contact locations.

\subsection{Experimental Setup}
To support the sensitivity and force-estimation evaluations, we collect a controlled normal-loading dataset using ten probes with different contact geometries. A CNC machine positions each probe at randomly selected locations on the sensor surface. At each location, four target force values are sampled randomly from the specified normal-force range. The probe is then pressed against the sensor at each target force while the tactile image stream and applied normal force are recorded synchronously.

The same loading apparatus and acquisition procedure are used for both GlowTact and the GelSight Mini. Normal force is measured using a calibrated load cell interfaced with an HX711 amplifier and analog-to-digital converter. This randomized sampling procedure produces measurements across a range of contact locations, applied forces, and probe geometries, reducing the likelihood that the resulting models depend on a fixed probing trajectory or a limited region of the sensor surface.

After cleaning and matching, the controlled-loading dataset contains
13{,}116 valid image--force samples for each sensor, collected using ten
probe geometries over a normal-force range of 0--20~N. The probe set spans
a range of contact shapes and surface geometries, enabling evaluation under
diverse contact conditions.

Tactile images are captured at $640 \times 480$ pixels with 30/25 Hz, while load-cell measurements are sampled at 80 Hz. The image and force streams are synchronized using timestamps and closest matching. Before each recording, we acquire unloaded frames and compute a reference image, $I_0$. For signal characterization, the tactile response is represented as the difference between each recorded image and this unloaded reference.

The controlled-loading dataset is used for two purposes. First, it is used to characterize the signal-to-noise ratio and contact-detection performance as functions of the applied normal force. Second, it is used to train and evaluate force-estimation models under variations in contact location, force magnitude, and probe geometry.

\begin{figure*}[t!h]
    \centering
    \includegraphics[width=0.9\textwidth]{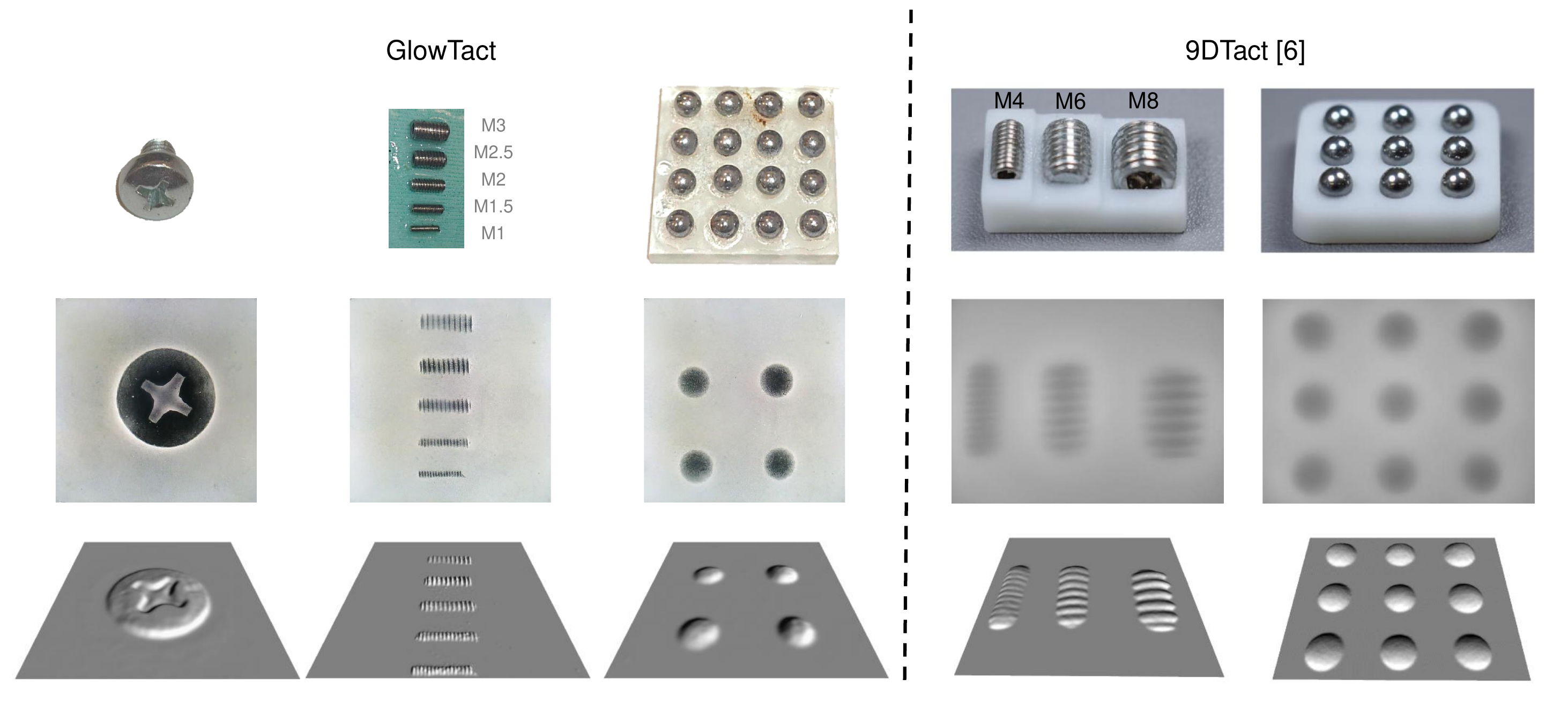}
    \vspace{-10 pt}
    \caption{\textbf{3D reconstruction of GlowTact, compared against 9DTact \cite{lin20239dtact}.}
    Top row: representative test objects. GlowTact is evaluated using an M5 Phillips screw head, threaded screws (M3, M2.5, M2, M1.5, and M1), and a steel-ball array. Representative 9DTact results \cite{lin20239dtact} are included for reference, with set screws (M4, M6, and M8) and a steel-ball array. 
    Middle row: corresponding tactile images.
    Bottom row: reconstructed contact geometries. Despite relying on a fundamentally different sensing mechanism, GlowTact retains sufficient spatial information to recover the overall contact geometry. }
    \label{fig:contact_reconstruction}
    \vspace{-10 pt}
\end{figure*}

\subsection{Contact Geometry}

GlowTact is designed to visualize pressure-induced optical coupling rather than reconstruct surface geometry. To evaluate the geometric information potentially available in GlowTact's tactile signal, we directly apply the reconstruction pipeline proposed for 9DTact~\cite{lin20239dtact}.

\autoref{fig:contact_reconstruction} shows representative reconstructions of
screw heads, threaded screws, and steel-ball arrays. Despite being developed
for a different sensing modality, the reconstruction pipeline recovers the
overall contact geometry from GlowTact observations, including structures such
as M1 threads (0.25~mm pitch). These results demonstrate that GlowTact maintains high pressure sensitivity while preserving fine geometric detail.

In \autoref{fig:contact_reconstruction} it is instructive to compare the outputs of GlowTact, shown on the left, with those of 9DTact, shown on the right.  Representative 9DTact responses are taken from
reference \cite{lin20239dtact}. GlowTact and 9DTact are similar in that they both provide a darkening signal which directly indicates pressure or depth. However, 9DTact's raw images (in the middle row) are much lower contrast and fuzzier than those of GlowTact. As a result, 9DTact's resolution hits its limit with an M4 screw (0.7 mm), while GlowTact is still providing sharp images with an M1 screw (0.25mm). Both 9DTact and GlowTact provide shape estimates that are qualitatively correct, although we have not made quantitative comparisons.

\subsection{Sensing Characteristics}

\begin{figure*}[t!h]
    \centering
    \includegraphics[width=1\textwidth]{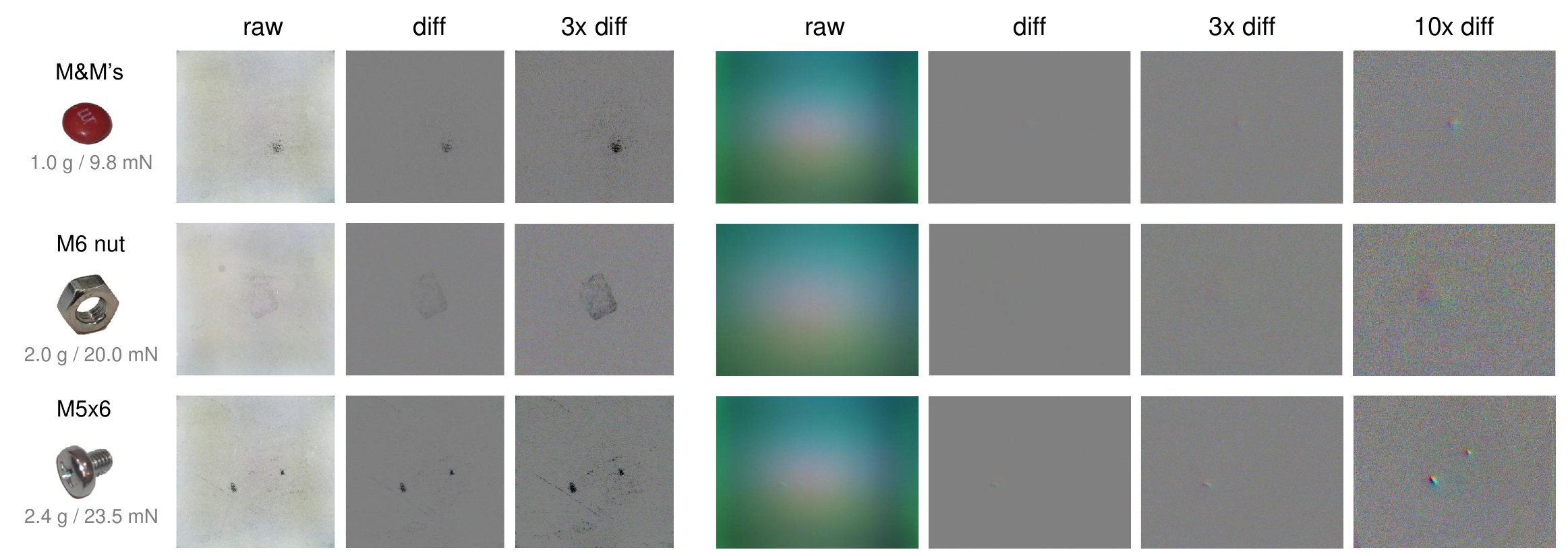}
    \caption{\textbf{Comparison of low-load contact sensitivity between GlowTact (left) and GelSight Mini (right).} Lightweight objects are gently placed on each sensor under their own weight without additional loading. For each object, the raw tactile image, the difference image, and a $3\times$ contrast-enhanced difference image are shown. GlowTact produces clear contact responses for all tested objects, whereas GelSight Mini exhibits weaker image changes under the same passive loading conditions. }
    \label{fig:light_object_test}
\end{figure*}

\paragraph{Passive Contact with Lightweight Objects.}
We first evaluate the ability of GlowTact to detect passive contacts under extremely light loading. Three lightweight objects, with masses ranging from 1.0 g to 2.4 g (corresponding to gravitational forces of 9.8--23.5 mN), are placed on GlowTact and GelSight Mini under their own weight, without any additional external loading. As shown in \autoref{fig:light_object_test}, GlowTact produces clear, spatially localized responses for all tested objects. GelSight Mini exhibits somewhat weaker image responses under the same conditions.  These qualitative observations indicate that GlowTact is highly responsive to contact formation even when the applied load is only on the order of tens of millinewtons.

\paragraph{Force Sensitivity}
Using the controlled-loading dataset described above, we compare the tactile
responses of GlowTact and GelSight Mini as functions of normal force. For each
frame, the response is computed as the normalized positive grayscale darkening
relative to the unloaded reference image. Noise statistics are estimated from
the unloaded frames, and the signal-to-noise ratio (SNR) is defined as the
median response divided by the unloaded standard deviation. Samples are grouped
into force bins, with response and SNR curves computed separately for each
probe and then equally averaged across probes. The minimum detectable force is
defined as the lowest force bin satisfying $\mathrm{SNR}\geq3$.

\begin{figure*}[!bht]
    \centering
    \includegraphics[width=0.8\textwidth]{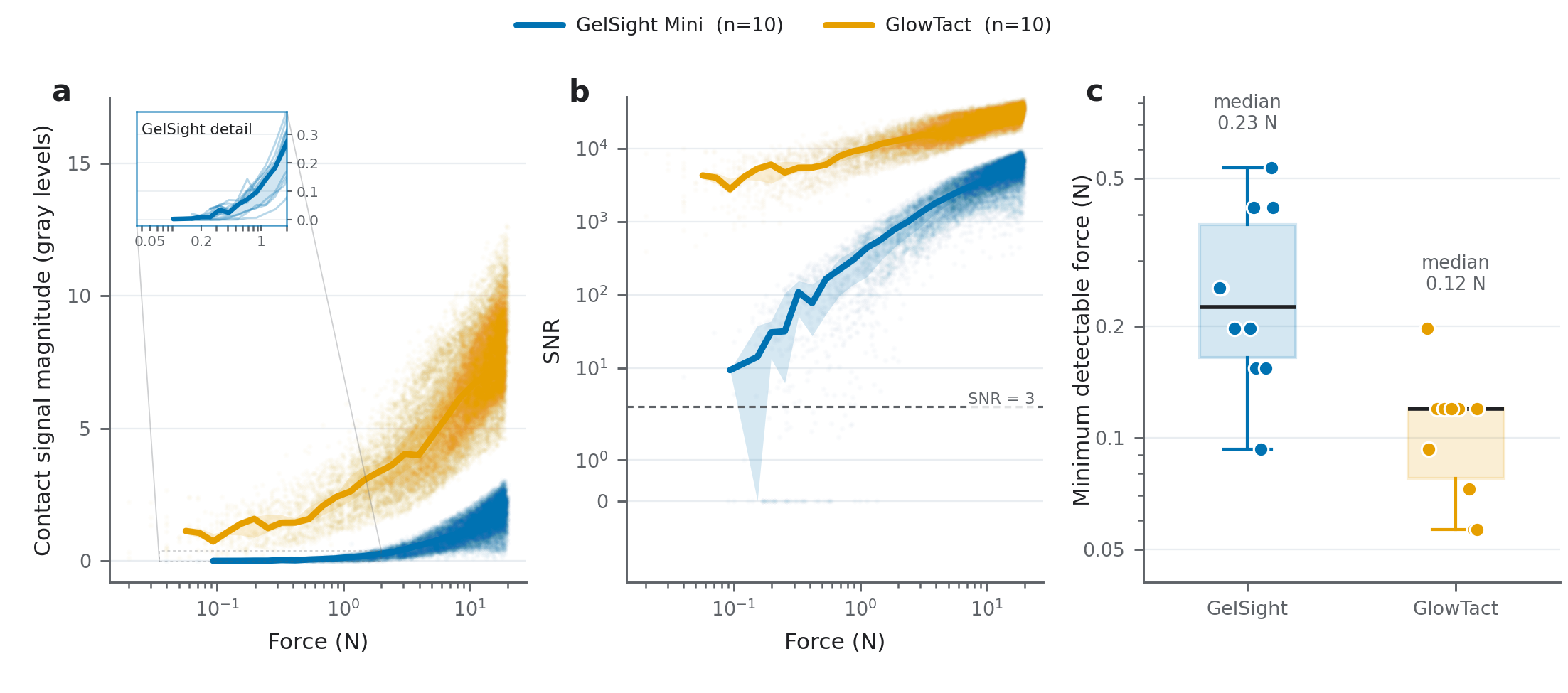}
\caption{
\textbf{Force sensitivity comparison between GlowTact and GelSight Mini.} (a) Mean tactile response as a function of normal force. (b) Signal-to-noise ratio (SNR) versus force; the dashed line denotes the detection threshold ($\mathrm{SNR}=3$). (c) Minimum detectable force for each probe geometry. GlowTact produces stronger low-force responses, higher SNR, and consistently lower detection thresholds than GelSight Mini.
}
    \label{fig:snr_test}
\end{figure*}

\subsection{Force Estimation}

Whereas the SNR analysis in \autoref{fig:snr_test} characterizes the
detectability of the raw tactile response, we next test whether this response
supports continuous normal-force estimation. Identical ResNet-18 regressors
are trained independently for GlowTact and GelSight Mini. Both models are
initialized with ImageNet-pretrained weights and optimized using the same
protocol.

After cleaning and matching, each sensor dataset contains
14{,}716 valid contact frames, comprising 13{,}116 frames from ten
controlled probe geometries and 1{,}600 frames from four everyday
objects (400 frames per object): a balloon, a light bulb, a pipe,
and a rope. The matched datasets span normal forces from 0 to 20~N. The same ten-probe subset is used for the SNR characterization in \autoref{fig:snr_test}.

For both sensors, 90\% of the spatial locations are used for training and
10\% for validation. All frames acquired at different depths of the same
indentation location are assigned to the same split, preventing frames from
a single loading trajectory from leaking across training and validation.
This protocol evaluates spatial generalization across multiple controlled
probe geometries and everyday objects.

The network input is the RGB difference between each contact frame and its
corresponding unloaded reference. During training, we apply random flips,
small translations, and multiplicative contrast scaling in
$[0.85,1.15]$.

The network jointly predicts normal force and contact area, with the latter
obtained automatically from the denoised contact mask. The auxiliary task
encourages the model to separate contact extent from optical contrast.
\autoref{tab:force_estimation} reports performance on held-out spatial
locations for both the controlled probes and four everyday objects.

\begin{table}[t]
\centering
\caption{Force-estimation errors at held-out spatial locations (N).}
\label{tab:force_estimation}

\small
\renewcommand{\arraystretch}{1.15}

\begin{tabular*}{\columnwidth}
{@{\extracolsep{\fill}}lcccc@{}}
\toprule
\multicolumn{5}{@{}l}{%
\textbf{a}\quad Controlled-probe validation} \\
\addlinespace[3pt]
&
\multirow{2}{*}{\shortstack{Overall\\RMSE}} &
\multicolumn{3}{c}{Force-bin MAE} \\
\cmidrule(lr){3-5}
Sensor &
&
0--0.5 &
0.5--2 &
2--20 \\
\midrule
GelSight Mini &
0.285 &
0.067 &
0.088 &
0.185 \\

GlowTact &
0.276 &
0.056 &
0.058 &
0.209 \\
\bottomrule
\end{tabular*}

\vspace{8pt}

\begin{tabular*}{\columnwidth}
{@{\extracolsep{\fill}}lcccc@{}}
\toprule
\multicolumn{5}{@{}l}{%
\textbf{b}\quad Everyday-object validation} \\
\addlinespace[3pt]
&
\multicolumn{2}{c}{GelSight Mini} &
\multicolumn{2}{c}{GlowTact} \\
\cmidrule(lr){2-3}
\cmidrule(lr){4-5}
Object &
MAE &
RMSE &
MAE &
RMSE \\
\midrule
Balloon &
0.134 &
0.161 &
0.087 &
0.112 \\

Light bulb &
0.151 &
0.189 &
0.093 &
0.116 \\

Pipe &
0.142 &
0.182 &
0.127 &
0.156 \\

Rope &
0.153 &
0.204 &
0.117 &
0.150 \\
\midrule
Macro average &
0.145 &
0.184 &
0.106 &
0.134 \\
\bottomrule
\end{tabular*}
\end{table}

Across the controlled probes, GlowTact consistently achieves lower
force-estimation errors than GelSight Mini over the evaluated force range.
The improvement is most pronounced at low forces, consistent with the
stronger optical response observed in \autoref{fig:snr_test}.

GlowTact also achieves a lower macro-average error across the four objects, indicating that its pressure-induced optical response provides a reliable force cue across variations in contact geometry and compliance. Together, these results show that the enhanced low-force sensitivity of GlowTact translates into more accurate quantitative force estimation.

\section{Conclusion}

We have introduced GlowTact, a novel technology for vision based tactile sensing. VBTS are attractive for humanoid fingertips because they can be fabricated in a wide range of shapes and sizes with highly compliant materials. Prior VBTS have a variety of limitations for humanoids: they may have limited sensitivity or resolution, or they may require computationally intensive processing, or they may offer challenging design issues in imaging or illumination due to the fingertip form factor.

GlowTact scores well on almost all dimensions. Its sensitivity and resolution are excellent: it can resolve the ridges on a fingerprint and can detect the weight of a 1 gm object. Compared to GelSight, it offers easier optical design, lower computational demands, and has a more durable skin. Through the use of a microtextured gel and an overlying sturdy black membrane, GlowTact provides video images that directly signal point by point pressure patterns. The direct utility of the camera output should improve the efficiency of robot learning systems that take tactile images as input.







\bibliographystyle{IEEEtran}
\bibliography{root}

\clearpage
\end{document}